\documentstyle[preprint,aps]{revtex}
\def\bqn{\begin{equation}}
\def\nqn{\end{equation}}
\def\bqnarr{\begin{eqnarray}}
\def\nqnarr{\end{eqnarray}}
\def\v#1{\left|\hspace{-0.5pt}\varepsilon_{#1}\rangle\right.}
\def\ket#1{\mid\hspace{-1.5pt}#1\rangle}
\def\bra#1{\langle#1\hspace{-1.5pt}\mid}
\def\H{\mbox{\bf H}}
\def\L{\mbox{\bf L}}
\begin{document}
\draft \preprint{} \title{Stabilizing the Richardson 
Algorithm by Controlling Chaos} \author{Song~He}
\address{Bell Laboratories, Lucent Technologies \\ Murray Hill, NJ 07974 \\ {\tt song@bell-lab.com} \\}
\date{May 9, 1996} 
\maketitle

\begin{abstract}
By viewing the operations of the Richardson purification algorithm as a discrete time 
dynamical process, we propose a method to overcome the instability of the algorithm by
controlling chaos.
We present theoretical analysis and numerical results on the behavior and performance of the stabilized algorithm. 
\end{abstract}
\vspace{0.2in}
%\pacs{PACS numbers: 02.70+d, 05.45+b, 89.80+h}

%\narrowtext
In 1950 Richardson\cite{Rich} proposed an algorithm to construct the eigenvectors
of a hermitian matrix when its eigenvalues are given. The basic idea of the method
is very simple: start from an arbitrary initial vector, and use the eigen equation to eliminate the unwanted components in the initial vector. Specifically, to compute the eigenvector $\v{k}$, we apply the operations according to the equation 
\bqn
\label{H}
\v{k}\sim \prod_{i\neq k} (\H-\varepsilon_i)\ket{\phi^{(0)}},
\nqn
where $\ket{\phi^{(0)}}$ is an arbitrary initial vector which has finite overlap with 
$\v{k}$ and $\{\varepsilon_i\}$ is the set of eigenvalues of the $N\times N$ hermitian matrix $\H$. 
If the operations can be done to
infinite precision, this procedure gives the desired eigenvector in $N-1$ iterations.
However, it is known that this algorithm is unstable\cite{Wilkinson}, {\it i.e.} \hspace{-2pt}an initial vector 
close to the desired eigenvector is driven away from it
under the operations of the algorithm. To see this, let us consider a vector 
\bqn
\ket{\phi} = \v{k} + \sum_{i\neq k} \delta_i \v{i},
\nqn
where $\delta_i$ are small compared to $1$. In the step of the algorithm when
$(\H-\varepsilon_j)$ is applied on $|\phi\rangle$, the resultant vector is proportional to
\bqn
\label{M1}
\ket{\phi^\prime} = \v{k} + \sum_{i\neq k} \delta_i^\prime \v{i},
\nqn
where 
\bqn
\label{M2}
\delta_i^{\prime}=\delta_i \frac{\varepsilon_i-\varepsilon_j}{\varepsilon_k-\varepsilon_j}.
\nqn
The above equations show that the fixed point $\v{k}$ of the map $(\H-\varepsilon_j)$ is
unstable. The exponents characterizing the instability can be as large as
$\ln W$, where $W$ is the natural band width defined by the difference between the
largest eigenvalue and the smallest eigenvalue devided by the smallest level spacing.
If the natural band width $W$ is large, then the
algorithm is highly unstable. It was proposed\cite{Wilkinson} that one may make the algorithm work better in special cases by replacing Eqn.(\ref{H}) by
\bqn
\label{HH}
\v{k}\sim \prod_{i\neq k} (\H-\varepsilon_i)^{n_i}\ket{\phi^{(0)}}
\nqn
with $n_i\geq 1$. However, this does not overcome the instability of the original 
algorithm. Experimentation
with this proposal showed that the scope of its applicability is very limited. Therefore the algorithm 
was concluded to have little practical use\cite{Wilkinson}. 

In this paper, we revisit the instability problem of the Richardson
algorithm. By viewing the operations
of the algorithm as a discrete time dynamical process and using ideas similar to controlling
chaos\cite{Con,Ott,Chaos}, we are able to devise a method to stabilize the algorithm. The key observation is that {\em by dynamically arranging the order of the operations in the product $\prod_{i\neq k} (\H-\varepsilon_i)^{n_i}\ket{\phi^{(0)}}$, we can use the instability itself to enhance the
amplitude of the desired eigenvector in the initial vector while suppressing those of the unwanted ones.}
In the rest of the paper, we first present the stabilized algorithm. Following that we present a theoretical analysis of the behavior of the stabilized algorithm. Finally we present numerical results on the behavior and the  the effectiveness of the stabilized algorithm: a randomly generated initial vector converges to the desired eigenvector exponentially on the average under the operations of the algorithm. 

To map the operations under the algorithm to a discrete time dynamical process, we
rewrite Eqn.(\ref{H}) as an iteration:
\bqn
\label{D}
\ket{\phi^{(n+1)}} = (\H - \varepsilon^{(n)}) \ket{\phi^{(n)}},
\nqn
where the sequence $\varepsilon^{(n)}\in\{\varepsilon_i,1\leq i \leq N, i\neq k\}$ defines the dynamics. 
As the discrete time $n$ increases, the 
initial state vector $\ket{\phi^{(0)}}$ of the dynamical system is transformed to $\ket{\phi^{(n)}}$ by the set of matrices $\{\H-\varepsilon_i: 0\leq i\leq N, i\neq k\}$. These matrices map the $N$-dimensional
unit sphere to itself if the vectors are normalized, which is assumed through out our paper. Eqn.'s (\ref{M1}) and (\ref{M2}) show that
the behavior of these maps around $\v{k}$ is very complex. There are large number of stable as well
as unstable directions. The behavior of the system is further complicated by rounding errors on a
finite precision computer. One of the consequences of a finite precision is that the operation in Eqn.(\ref{D}) becomes nonlinear and the system becomes chaotic. To see this, consider the operation of $\H-\varepsilon_l$ on a vector
$\ket{\phi}$ on a finite precision computer. To be more precise, we define a function 
\bqn
\Re_{\varepsilon_l,\varsigma}(\ket{\phi})\equiv(\H-\varepsilon_l)\ket{\phi},
\nqn
where $\varsigma$ is the precision of the computer, {\it i.e.} \hspace{-2pt}for any $x\in(-\varsigma,\varsigma)$, the computer gives $1+x=1$. Let $\ket{\phi}=\v{i}+\gamma\v{j}$ and $\gamma$ be sufficiently small. 
We have on the one hand
\bqn
\label{NL1}
\Re_{\varepsilon_l,\varsigma}(\ket{\phi})=\Re_{\varepsilon_l,\varsigma}(\v{i})=(\varepsilon_i-\varepsilon_l)\v{i}.
\nqn
On the other hand,
\bqn
\label{NL2}
\Re_{\varepsilon_l,\varsigma}(\v{i})+\gamma\Re_{\varepsilon_l,\varsigma}(\v{j})=(\varepsilon_i-\varepsilon_l)\{\v{i}+\gamma^{\prime}\v{j}\},
\nqn
where 
\bqn
\gamma^{\prime}=\gamma\frac{\varepsilon_j-\varepsilon_l}{\varepsilon_i-\varepsilon_l}.
\nqn
Clearly if the natural band width $W$ is large enough, the term proportional to $\gamma^{\prime}$ on the right hand side of Eqn.(\ref{NL2}) may not be negligible. Therefore the function $\Re_{\varepsilon_l,\varsigma}(\ket{\phi})$ is {\em nonlinear}. We emphasize that the nonlinearity
of $\Re_{\varepsilon_l,\varsigma}(\ket{\phi})$ also makes the order of the operations in Eqn.(\ref{H}) or Eqn.({\ref{HH}) important. In Fig.\ref{LYA},
we show the typical behavior of the system described by Eqn.(\ref{D}). The lower panel shows the 
separation $\zeta=\|\ket{\phi_1}-\ket{\phi_2}\|$ between two initially neighboring vectors as a function of the discrete time. The initial separation between these two vectors is about $10^{-16}$. The upper panel shows the computed Lyapunov exponent $\lambda$, which converges to about $0.23$.  
The matrix used to obtain these results is a $4096\times 4096$ tridiagonal matrix, which is more precisely defined
later in the paper. We choose $k$ to be $1$. The sequence $\{\varepsilon^{(n)}\}$ which determines the dynamics is periodic and is given by $\varepsilon^{(n)}=\varepsilon_{P_j}$, where $j=n\bmod(N-1)$ and $P$ is a random permutation of the $N-1$ indices $\{2,\cdot\cdot\cdot,N\}$. The irregular behavior of $\zeta$ in the lower panel
of the figure is not due to any randomness. It is due to the deterministic chaotic dynamics defined
by Eqn.(\ref{D}) on a finite precision computer. The task of stabilizing the algorithm is to determine
$\varepsilon^{(n)}$ dynamically from the set of {\em discrete} values $\{\varepsilon_i,i\neq k\}$ so that $\ket{\phi^{(n)}}$ converges to $\v{k}$ as $n$ increases. The ideas of
controlling chaos come into the construction of the sequence $\{\varepsilon^{(n)}\}$. It is interesting 
to contrast our case with that of the usual problem of controlling chaos\cite{Ott,Chaos} where a set of {\em continuous} parameters is adjusted in dragging the system back to the desired periodic orbit.

The iterative procedure to compute the eigenvector $\v{k}$ using the stabilized algorithm is the following.
\newcounter{a}
\begin{list}
{A-\arabic{a}}{\usecounter{a}\setlength{\rightmargin}{\leftmargin}}
\item Generate a normalized random initial vector $\ket{\phi^{(0)}}$ and an random array $\{a_i^{(0)}:0<a_i^{(0)}<1\}$.
\item In the $n$'th step of the algorithm, the parameter $\varepsilon^{(n)}$ is determined by
$\varepsilon^{(n)}=\varepsilon_j$, where $j$ is such that $a_j^{(n)}={\rm max}_{i\neq k}\{a_i^{(n)}\}$. The {\em normalized} vectors $\ket{\phi^{(n)}}$ and the array $\{a_i^{(n)}\}$ are updated according to the equations                                                                                                            \bqn
\ket{\phi^{(n+1)}} = (\H-\varepsilon^{(n)})\ket{\phi^{(n)}},
\nqn
\bqn
a_i^{(n+1)}=\left\{\begin{array}{ll}
\frac{\bar{\delta}}{\Delta} & i=j, \\
\frac{a_i^{(n)}\mid\varepsilon_i-\varepsilon^{(n)}\mid}{\Delta} & i\neq j,
\end{array}
\right.
\nqn
where $\Delta$ is the smallest level spacing, $\bar{\delta}<\Delta$ is a small number comparable to the 
accuracy $\delta$ (defined in Eqn.(\ref{DD})) to which the eigen equation is satisfied. 
\item Terminate the iteration if the error parameter
$\bar{\sigma}=\{\frac{1}{N}\parallel(\H-\varepsilon_k)\ket{\phi^{(n)}}\parallel^2\}^{\frac{1}{2}}$ is smaller than required.
\end{list}

Let $M$ be the total number of iterations and $m_i$ be the number of occurrences of $\varepsilon_i$ in the sequence $\{\varepsilon^{(n)}\}$.
Clearly $m_k=0$. Less obviously $m_i\propto M$ as $M\longrightarrow\infty$. The latter relation comes about for the following reason: A newly eliminated element in the array $\{a_i^{(n)}\}$ is set to a small residual value $\bar{\delta}/\Delta$ in the iteration. It grows exponentially in the later iterations and is chosen to be eliminated again by the algorithm.

To analyze the behavior of the algorithm, we need to obtain the
accumulated effect of the iterations on the initial vector. For that purpose, we write
\bqn
\ket{\phi^{(0)}}=\sum_i c_i^{(0)}\v{i},
\nqn
where $\v{i}$ are the eigenvectors of the matrix $\H$. On a finite precision computer, the
eigen equation is only satisfied to about the machine precision, {\it i.e.} \hspace{-2pt}we have
\bqn
\label{EI}
(\H-\varepsilon_i)\v{i}=\sum_j R_{i,j}\v{j},
\nqn
where $R_{i,j}$ is the same order as the machine precision $\varsigma$. 
The resultant vector is
\bqn
\ket{\phi^{(M)}}\propto\prod_n(\H-\varepsilon^{(n)})\cdot\sum_i c_i^{(0)}\v{i}=\sum_i c_i^{(M)}\v{i},
\nqn
where 
\bqn
c_i^{(M)}=c_i^{(0)}\prod_n(\varepsilon_i-\varepsilon^{(n)})=c_i^{(0)}\prod_{j\neq i}
(\varepsilon_i-\varepsilon_j)^{m_j} R_{i,i}^{m_i},
\nqn
and the terms higher order in $\varsigma$ have been ignored.
To show that $\ket{\phi^{(M)}}$ indeed converges to $\v{k}$, we compute the ratio
\begin{eqnarray}
\label{R}
\left|\frac{c_i^{(M)}}{c_k^{(M)}}\right| &\leq& \left|\frac{c_i^{(0)}}{c_k^{(0)}}\right|\prod_{j\neq i,k}
\left|\frac{\varepsilon_i-\varepsilon_j}{\varepsilon_k-\varepsilon_j}\right|^{m_j}\left|\frac{\delta}
{\varepsilon_k-\varepsilon_i}\right|^{m_i}\\
                                    &\equiv& \left|\frac{c_i^{(0)}}{c_k^{(0)}}\right| r_{i,k}^{(M)},
\end{eqnarray}
where the parameter $\delta$ is defined by 
\bqn
\label{DD}
\delta=\max_{i\neq k}\{\left|R_{i,i}\right|\}.
\nqn
We remark that: (1) If $\left|\frac{\delta}
{\varepsilon_k-\varepsilon_i}\right|\ll 1$, which usually is the case in practical applications, 
the vanishing ratio $r_{i,k}^{(M)}$ is dominated by
$\left|\frac{\delta}{\varepsilon_k-\varepsilon_i}\right|^{m_i}$. The initial values of $\{c_i^{(0)}\}$
do not affect the convergence of the algorithm. (2) The motivation of introducing the array $\{a_i^{(0)}\}$ in the stabilized algorithm is to replace the {\it a priori} unknown coefficients $\left|c_i^{(0)}\right|$ by $a_i^{(0)}$. The previous comment justifies our so doing.  

To demonstrate the practical effectiveness and to illustrate the numerical behavior of our algorithm, we show two sets of numerical results. In the first example, we deal with computing the eigenvectors of of a tridiagonal matrix
after its eigenvalues have been obtained by the usual implicit QL method\cite{Wilkinson,Eispack}.
In the second example, we deal with the problem of computing the basis vectors in the decomposition
of a direct product of irreducible representations of the $SU(2)$ algebra. In this case, the algorithm 
is especially effective because the eigenvalues
of the matrix involved, the total angular momentum operator, are known exactly due to the algebraic
properties of $SU(2)$.

In the first example, we demonstrate the behavior and the performance of the stabilized algorithm 
by computing the eigenvectors of a tridiagonal matrix of dimension $N=4096$. The diagonal terms $H_{i,i}\in [-1,1]$ are randomly generated with a uniform distribution. 
Its next diagonal terms are all $1$.
The eigenvalues of the matrix are obtained by the standard implicit QL procedure. For the
particular matrix we used, the smallest level spacing $\Delta$ equals $1.67\times 10^{-6}$ and the natural band 
width 
$W$ equals $2.7\times 10^6$. The parameter $\bar{\delta}$ should be the same order as the machine precision, which is usually 
$10^{-13}$ with double precision. In our calculations, we choose it to be $\bar{\delta}=10^{-10}$. We will comment
on the effect of the exact magnitude of $\bar{\delta}$ on the behavior of the algorithm later in the paper. 

In Fig.\ref{R0} and Fig.\ref{R1}, we show respectively the error parameter $\sigma$ defined by
\bqn
\sigma=\{\frac{1}{N}\|\ket{\phi^{(n)}}-\v{k}\|^2\}^{\frac{1}{2}}
\nqn
as a function of $n$ for $k=1$, the lowest eigenvalue, and $k=3097$, the eigenvalue next to the smallest level spacing $\Delta$. For the eigenvector $\v{k}$ in the above equation, we use the converged eigenvector obtained from our stabilized algorithm according to A-1, A-2, and A-3.
For comparison, the results for $\sigma$ calculated from a naive implementation of algorithm without controlling chaos are also shown as dashed lines in the figures. In the naive implementation
of the algorithm, the unwanted components of the initial vector are eliminated periodically instead of according to the procedure outlined in A-1, A-2, and A-3. The effectiveness of the stabilized algorithm is evident: the error parameter $\sigma$ decays exponentially on the average as a function of $n$ after an initial transient period. In particular, for $k=1$, we see that the initial vector converges to the desired eigenvector within machine precision in about $M=300$ iterations. This is much fewer than the number of iterations $N-1=4095$ required by an implementation of the original algorithm with infinite precision. This shows that by dynamically arranging the order
of the operations, we can indeed use the instability itself to enhance the desired component in the initial
vector while suppressing the unwanted ones. For $k=3097$, the initial vector manages to converge to the desired eigenvector within machine precision in about $1.5\times 10^{4}\sim 4N$ iterations. We find that this eigenvector is the most difficult to compute because its eigenvalue is next to the smallest level spacing $\Delta$. This is similar to other algorithms\cite{Wilkinson} for
computing eigenvectors. We also notice that $\sigma$ exhibits large fluctuations as a function of
$n$, even though its magnitude always remains at around $10^{-10}$. Detailed analysis of the operations
of the algorithm indicates that these fluctuations are due to the instability of the original algorithm.
However, we want to emphasize that these fluctuations do not affect the computation of the desired eigenvector since we can always terminate the iteration once $\sigma$ is smaller than required. 

In Fig.\ref{R2}, we show the quantity $r_{i,k}^{(M)}$ defined in Eqn.(\ref{R}) as a function of $i$ for $k=3097$. In the inset, we show $m_i$ as a function of $i$. The similarity between the curves in the figure
show that $r_{i,j}^{(M)}$ is indeed dominated by the contribution from $\left|\frac{\delta}{\varepsilon_k-\varepsilon_i}\right|^{m_i}$, as asserted above in the analysis of the algorithm. The largest magnitude of $r_{i,k}^{(M)}$ is very small, about $10^{-25}$. This guarantees the convergence to the desired eigenvector for almost any initial vector $\ket{\phi^{(0)}}$. We also find that the relative magnitude of $m_i$ depends on the local level spacing around the
eigenvalue $\varepsilon_i$. Our calculations indicate that $m_i$ is large (small) when the local level spacing is large (small).  

We have experimented with our algorithm on variety of matrices with different eigen spectra structure. 
Our experience indicates that convergence is maintained even when $\delta/\Delta$ is comparable to $1$. 
We find that the number of iterations $M$ needed for the
algorithm to compute a converged eigenvector of the low-lying eigenvalues within machine precision does not increase much as $N$ increases
 for any given
class of matrices. This means that for these eigenvectors our algorithm converges in about $\alpha N^2$ operation, where $\alpha$ depends
on the eigen structure of the matrices and is a very slowly varying function of $N$. In the case of tridiagonal matrices with random
diagonal elements and for the eigenvector of the lowest eigenvalue, our calculations on matrices of
dimensions from $512$ to $131072$ indicate that $\alpha$ is basically a constant around $300$. For a typical eigenvector, the algorithm converges in about $M=\beta N$ iterations, {\it i.e.} \hspace{-2pt}$\beta N^3$ operations, where $\beta$ is a number comparable to $1$. The
exactly value of $\beta$ depends on the local level spacing. We find that the parameter $\bar{\delta}$ used in the algorithm has effects similar to that of the temperature in an optimization problem by simulated annealing. When $\bar{\delta}/\Delta$ is small, we find that the vector $\ket{\phi^{(n)}}$ may be trapped around a particular vector for a long time before it manages to escape. On the other hand, if $\bar{\delta}/\Delta$ is large, we find that it takes longer for the initial vector to converge and the error parameter $\sigma$ fluctuates significantly as a function of $n$. Our experience also indicates that refreshing (replacing $a_i^{(n)}$ by its reciprocal, for example) the array $\{a_i^{(n)}\}$ periodically when $\sigma$ has become small, $10^{-5}$ for example, helps to speed up the convergence. We believe this eliminates the bias built into 
the relative values of $\{a_i^{(n)}\}$ according to Eqn.(\ref{R}) during the iterations. In comparison to 
other traditional methods of computing eigenvectors, the stabilized algorithm has a number of advantages.
Here we want to compare it to the inverse iteration method, the best traditional method of computing eigenvectors when the eigenvalues are known\cite{Wilkinson}. It typically takes about $N^3$ operations for
inverse iteration to compute an eigenvector. This is comparable to the number of operations needed by our algorithm in the worst cases. However, the inverse iteration method needs more computer memory because it needs
to store the inverse of the matrix $\H-\varepsilon_k$. If the matrix $\H$ is unstructurally sparse, it is
clear that our algorithm can take advantage of the sparsity but the inverse iteration method cannot because
the inverse of $\H-\varepsilon_k$ may not be a sparse matrix.

In the second example, we use the algorithm to construct the basis vectors in the decomposition
of a direct product of irreducible representations of the algebra $SU(2)$. The specific case
we studied is to construct the basis vectors of the irreducible representations contained in the
product $\underbrace{S\otimes S\otimes\cdot\cdot\cdot\otimes S}_N$. This is a generalization of the usual Clebsch-Gordan coefficients used to couple two angular momenta together.
In our example, what we want to do physically is to construct the eigenstates of the total angular momentum operator $\L$ of $N$ spinless electrons with orbital 
angular momentum $S$. This is very useful in block diagonalizing the Hamiltonian of a many 
electron system\cite{He}. To be more specific, we choose $N=8$ and $S=\frac{21}{2}$. From the algebraic
properties of $SU(2)$, we know that $\underbrace{\frac{21}{2}\otimes\frac{21}{2}\otimes\cdot\cdot\cdot\otimes\frac{21}{2}}_{8}=0\oplus 1\oplus\cdot\cdot\cdot\oplus 56$ for identical fermions. Therefore the eigenvalues of the total angular
momentum operator $\L^2$ are {\em known} exactly: $\{\varepsilon_l=l(l+1), 0\leq l \leq 56\}$. When the $z$ component of the angular momentum is constrained to be $\L_z=0$, the 
dimension of the matrix $\L^2$ is $D=8512$. This is much larger than the number of distinct eigenvalues. Therefore the eigenvalues are highly degenerate. To construct the basis vectors of the eigen subspace
$\Omega_l$ of dimension $D_l$, we simply start from $D_l$ randomly generated initial vectors $\{\ket{\phi^{(0)}_{l}}, l=1,2,\cdot\cdot\cdot, D_l\}$ and apply our stabilized algorithm. It is easy
to show that the final eigenvectors are linearly independent with probability 1.

In Fig.(\ref{L0}), we show the error parameter $\sigma$ as a function of $n$ for $l=0$. Since the
eigenvalues are massively degenerate, we have to modify the definition of $\sigma$ to
\bqn
\sigma=\{\frac{1}{N}\|\ket{\phi^{(n)}}-\sum_{i=1}^{D_l}\ket{e_l^{(i)}}\bra{e_l^{(i)}}\phi^{(n)}\rangle\|^2\}^{\frac{1}{2}},
\nqn
where $\{\ket{e_l^{(i)}}\}$ is the known basis set of the eigen subspace $\Omega_l$.
The data shown in Fig.(\ref{L0}) are obtained for $l=0$. The dimension of the eigen subspace is $D_0=31$. The dashed line shows $\sigma$ obtained from a naive implementation of the original algorithm defined
in previous discussions. We see that a randomly generated initial vector converges to
the desired eigenvector within machine precision in about $M=75$ iterations. We note that fluctuations in
$\sigma$ are also present, similar to the first example. We have studied the behavior of the algorithm
with other values of $l$, $N$, and $S$ where the dimension of the matrix $\L^2$ can be as large as $10^6$. 
We find that in all cases, the initial vector converges to the desired
eigenvector within machine precision in a number of iterations fewer than twice of the number of distinct 
eigenvalues of the operator $\L^2$. We conjecture that this is true for all values of $l$, $N$, and $S$. 
A useful point to note in speeding up the construction of the
basis vectors is that we may use the same sequence $\{\varepsilon^{(n)}\}$ for the construction of all of the basis vectors. This avoids the computation efforts in constructing $\{\varepsilon^{(n)}\}$ for each
of the initial vectors. One can show that the basis vectors generated this way are linearly independent with probability 1 due to the instability. 

We expect the stabilized algorithm to have similar advantages for other compact Lie algebras. The
reasons are mainly 
twofold. (1) The algebraic properties of the Lie algebra involved usually determine the eigen
spectra of the operators involved. The eigenvalues are known exactly and do not need to be computed.
(2) The dimensions of the irreducible representations are usually large. This means that the eigen
spectra of the operators involved are highly degenerate. Therefore the number of
distinct eigenvalues and the natural band widths are 
small. Because of these features, we find the stabilized algorithm to be a very powerful tool in fully
utilizing the symmetry of the physical system concerned\cite{He}.

Before concluding, we remark that the algorithm should work for general non-hermitian matrices as well. Depending on whether the desired eigenvector is a {\em left} or {\em right} eigenvector, we
need to modify the operations of the matrix on the vector $\ket{\phi^{(n)}}$ to {\em left} or {\em right} multiplications. As a general remark, we find the idea of viewing the operations of an 
algorithm as a dynamical process very intriguing. It seems reasonable to believe that the explorations
in this direction should be interesting and fruitful.

In summary, we have discussed a method to stabilize the Richardson purification 
algorithm using controlling chaos. We have presented theoretical analysis and numerical results to
illustrate the behavior and the effectiveness of the stabilized algorithm. We find that by
dynamically arranging the order of operations in the originally unstable algorithm, we are able to
use the instability to enhance the desired components of a initial vector while suppressing the unwanted ones so that the initial vector converges to the desired eigenvector exponentially. 

\noindent Acknowledgments: The author would like to thank P.~B.~Littlewood for reading the manuscript
and making helpful suggestions. He would also like to thank Ming-Zhou Ding for helpful discussions.

\begin{figure}
\caption{Generic behavior of the system. Upper panel: Lyapunov exponent. Lower panel: separation
between two initially neighboring vectors.}
\label{LYA}
\end{figure}
\begin{figure}
\caption{The error parameter $\sigma$ as a function of the number of iterations for the lowest eigenvalue. 
Solid line: with controlling chaos. Dashed line: without controlling chaos.}
\label{R0}
\end{figure}
\begin{figure}
\caption{The error parameter $\sigma$ as a function of the number of iterations for the eigenvalue next to the 
smallest level spacing $\Delta$. Solid line: with controlling chaos. Dashed line: without controlling chaos.}
\label{R1}
\end{figure}
\begin{figure}
\caption{$\log_{10} r_{i,k}^{(M)}$ as a function of the eigenvalue index $i$ for $k=3096$. 
Inset shows $m_i$ as a function of $i$.}
\label{R2}
\end{figure}
\begin{figure}
\caption{The error parameter $\sigma$ as a function of the number of iterations for $l=0$. 
Solid line: with controlling chaos. Dashed line: without controlling chaos.}
\label{L0}
\end{figure}


\begin{references}
\bibitem{Rich} L.~F.~Richardson, {\it Phil. Trans. Roy. Soc.} A242, 439(1950).
\bibitem{Wilkinson} J.~H.~Wilkinson, ``{\it The Algebraic Eigenvalue Problem}'', Clarendon Press,
Oxford, 1965.
\bibitem{Con} A.~Hubler, {\it Helv. Phys. Acta} {\bf 62}, 343(1989); T.~B.~Fowler, {\it
IEEE Trans. Autom. Control} {\bf 34}, 201,(1989).
\bibitem{Ott} E.~Ott, Celso~Grebogi, and J.~A.~Yorke, {\it Phys. Rev. Lett.} 64, 1196(1990).
\bibitem{Chaos} D. Auerbach, C. Grebogi, E. Ott and J. A. Yorke, {\it Phys. Rev. Lett.} 69, 3479(1992); U.~Dressler and G.~Nitsche, {\it Phys. Rev. Lett.} 68, 1(1991); Hu Guang and Qu~Zhilin,
{\it Phys. Rev. Lett.} 72, 68(1994); R. Roy, T. W. Murphy, Jr., T. D. Maier, and Z. Gills,
{\it Phys. Rev. Lett.} 68, 1259(1992).
\bibitem{Eispack} B.~T.~Smith, {\it et al.} ``{\it Matrix Eigensystem Routines--Eispack guide}'', Springer-Verlag, New York,
1976.
\bibitem{He} Song He, S.~H.~Simon, and B.~I.~Halperin, {\it Phys. Rev.} {\bf B}50, 1823(1994).

\end{references}
\end{document}